\documentclass[runningheads]{llncs}

\usepackage{eccv}

\usepackage{eccvabbrv}

\usepackage{graphicx}
\usepackage{booktabs}

\usepackage[accsupp]{axessibility}  

\usepackage{hyperref}

\usepackage{orcidlink}
\usepackage{textcomp}

\begin{document}

\title{
From Synchrony to Sequence: Exo-to-Ego Generation via Interpolation}
\titlerunning{Syn2Seq-Forcing}

\author{Mohammad Mahdi\inst{1}\thanks{\email{firstname.lastname@insait.ai}} \quad Nedko Savov\inst{1} \quad
Danda Pani Paudel\inst{1}  \quad Luc Van Gool\inst{1}}

\authorrunning{Mahdi et al.}

\institute{$^1$INSAIT, Sofia University “St. Kliment Ohridski”}

\maketitle

\begin{abstract}
Exo-to-Ego video generation aims to synthesize a first-person video from a synchronized third-person view and corresponding camera poses. While paired supervision is available, synchronized exo-ego data inherently introduces substantial spatio-temporal and geometric discontinuities, violating the smooth-motion assumptions of standard video generation benchmarks. We identify this synchronization-induced jump as the central challenge and propose {Syn2Seq-Forcing}, a sequential formulation that interpolates between the source and target videos to form a single continuous signal. By reframing Exo2Ego as sequential signal modeling rather than a conventional condition–output task, our approach enables diffusion-based sequence models, e.g. Diffusion Forcing Transformers (DFoT), to capture coherent transitions across frames more effectively. Empirically, we show that interpolating only the videos, without performing pose interpolation already produces significant improvements, emphasizing that the dominant difficulty arises from spatio-temporal discontinuities. Beyond immediate performance gains, this formulation establishes a general and flexible framework capable of unifying both Exo2Ego and Ego2Exo generation within a single continuous sequence model, providing a principled foundation for future research in cross-view video synthesis. The code will be released at \url{https://github.com/insait-institute/Syn2Seq}.

\end{abstract}
\begin{figure}[h]
  \centering
 \includegraphics[width=1.0\linewidth]{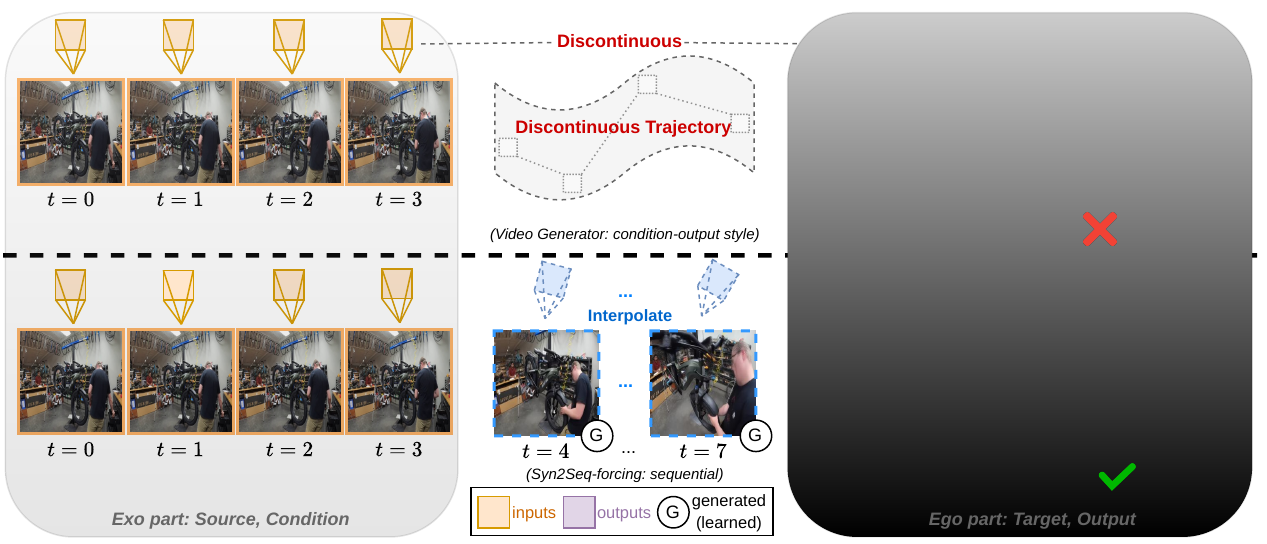}
  \caption{Top: Standard video generators struggle with discontinuous camera poses and missing transitions from exo to ego. Bottom: Syn2Seq-Forcing (ours) interpolates between views and camera poses to form a single continuous sequence, enabling smooth exo2ego generation; the same framework naturally extends to ego2exo. }
  \label{fig:teaser}
  \vspace{-0.1in}
\end{figure}

\section{Introduction}
\label{sec:intro}


Given temporally synchronized exocentric (third-person) and egocentric (first-person) videos of the same scene, denoted as the source video $x$ and the target video $g$, along with their corresponding per-frame camera poses $p = [p_x, p_g]$, the Exo2Ego generation task aims to learn a parametric mapping $f_\theta$ that synthesizes the target egocentric observation conditioned on the available source view and camera geometry:

\[
\min_{\theta} \; \| f_\theta(x, p) - g \|.
\]

Exo2Ego generation is practically significant in applications such as robotics, augmented reality, and virtual reality~\cite{li2024egoexo,xu2025egoexo,he2025egoexobench,mahdi2025exo2egosyn,huang2024egoexolearn,liu2020exocentric}, where systems must reconstruct or predict a first-person experience from third-person observations to enable accurate perception, interaction, and immersive feedback. The problem is inherently challenging: it not only requires transferring viewpoint information but also necessitates capturing fine-grained spatial and temporal cues, including body dynamics, hand-object interactions, and scene elements that may be only partially observable from the exocentric perspective. Moreover, maintaining temporal coherence and visual fidelity across frames further increases the complexity of the task.

While temporal synchronization provides paired supervision, it also introduces two fundamental challenges that make Exo2Ego generation qualitatively different from standard video generation. First, from a frame-coherence perspective, the viewpoint gap between $x$ and $g$ is often substantial, resulting in a large spatio-temporal discontinuity between the end of the source video and the beginning of the target video. Second, from a geometric perspective, the corresponding camera poses can exhibit significant discontinuities: the configuration of $p_x$ at the end of the source sequence may differ sharply from $p_g$ at the start of the target sequence. 

In contrast, most conventional video generation benchmarks~\cite{bai2025recammaster,yu2025trajectorycrafter} assume smooth motion and continuous trajectories. Consequently, models trained under these assumptions, e.g.,~\cite{bai2025recammaster}, often struggle to handle the abrupt spatial and geometric transitions inherent in synchronized exo-ego video pairs. These challenges highlight the need for methods that explicitly account for both \textit{viewpoint jumps} and \textit{pose discontinuities} to produce temporally coherent and visually accurate egocentric outputs.

Prior work has largely followed a condition–output paradigm, exploring different strategies to condition on $x$ and $p$~\cite{liu2024exocentric,luo2024put} within various architectures \cite{mahdi2025exo2egosyn}, or leveraging multiple exocentric views to enrich spatial cues \cite{li2024egoexo}. In this paper, however, we take a step back to address the fundamental challenge introduced by synchronization. Our key idea is to replace the traditional \textit{synchronized but discontinuous} formulation with a \textit{sequential} one: we interpolate between $x$ and $g$ to generate a single continuous signal that effectively bridges the spatio-temporal and geometric gaps, as illustrated in Fig.~\ref{fig:teaser}. 

This transformation, moving from condition-output modeling to sequential signal modeling, not only produces measurable improvements in performance but also opens up new avenues for flexible and controllable video generation. By treating the source, interpolated, and target sequences as a unified signal, where past frames condition future frames, our approach enables the network to capture smooth transitions, maintain temporal coherence, and better handle abrupt viewpoint and pose changes inherent in exo–ego video pairs.

Exo2Ego generation requires strong depth understanding, which prior work typically addresses using explicit 3D priors. In contrast, we adopt interpolation as a simple proxy for depth reasoning. Rather than explicitly estimating geometry, we guide the model toward the target viewpoint through intermediate frames—essentially “\textit{if the depth is unknown, reach the target via interpolation}.”

We demonstrate that even interpolating only the video frames without performing interpolation on the camera poses, already leads to substantial gains, emphasizing that the primary challenge in Exo2Ego generation stems from the spatio-temporal discontinuities introduced by synchronization. Beyond these performance improvements, the sequential formulation offers significant conceptual flexibility: in principle, it could be extended to generate longer sequences, such as transitioning from egocentric back to exocentric views, thereby enabling both Exo2Ego and Ego2Exo generation within a single unified framework. Although we only briefly explore this extended flexibility in the present work, the results highlight the broader potential of our approach as a general framework for exo/ego-centric video generation.

\noindent{\textbf{Contributions.}}
\begin{enumerate}
    \item We study and identify synchronization-induced discontinuities as the core obstacle in Exo2Ego video generation.
    \item We resolve this obstacle by reframing the task as sequential modeling with interpolation, yielding significant performance gains even without pose interpolation.
    \item We introduce a general-purpose paradigm that can unify both Exo2Ego and Ego2Exo generation within a single continuous sequence model.
\end{enumerate}

\section{Related Works}
\label{sec:rw}

\noindent{\textbf{Diffusion-based Video Generation.}}

Recent advances in diffusion models and large-scale datasets have rapidly improved the quality and diversity of video generation~\cite{rombach2022high,huang2024learning,liu2025control,liu2025diverse,pan2025earthsynth,blattmann2023align,ho2022imagen,ho2022video,zhang2025show,zhou2022magicvideo,peebles2023scalable,yang2024cogvideox}. Representative systems such as Tune-A-Video~\cite{wu2023tune}, Video Diffusion Models~\cite{ho2022video}, Stable Video Diffusion~\cite{blattmann2023stable}, AnimateDiff~\cite{guo2023animatediff}, LTX~2~\cite{hacohen2026ltx}, HunyanVideo~\cite{kong2024hunyuanvideo} and WAN2.2~\cite{wan2025wan} demonstrate the ability of diffusion models to generate temporally coherent and visually realistic videos. 
Recently, long-horizon video generation has been advanced by autoregressive and diffusion-based approaches, including Diffusion Forcing Transformers~\cite{song2025history}, SelfForcing~\cite{huang2025self}, and SelfForcing++~\cite{cui2025self}, which enable stable generation over extended sequences.

To offer fine-grained control beyond an initial text prompt, recent work has introduced mechanisms for explicit camera control in video generation. Methods such as CameraCtrl~\cite{he2024cameractrl}, MotionCtrl~\cite{wang2024motionctrl}, Direct-a-Video~\cite{yang2024direct}, and CamTrol~\cite{hou2024training} enable controllable camera trajectories during generation. ReCamMaster~\cite{bai2025recammaster} achieve fine-grained camera control by directly manipulating temporal attention within large video generators. World models enable interactive environments with real-time camera control:  WorldPlay~\cite{sun2025worldplay}, Genie~3~\cite{parker2025genie}, and LingBot~\cite{team2026advancing}. Long-horizon video generators, such as Diffusion Forcing Transformers~\cite{song2025history} and MotionStream~\cite{shin2025motionstream} have been demonstrated to follow camera controls as well. SPMem \cite{wu2025video}, VMem \cite{li2025vmem}, Gen3C \cite{ren2025gen3c}, and TrajectoryCrafter \cite{yu2025trajectorycrafter} maintain 3D representations during video generation, to produce consistent content throughout viewpoints.

These camera-controlled video generation methods synthesize videos by following a continuous camera motion trajectory during generation. However, they do not address the task of transforming the viewpoint of an existing video, leaving cross-view camera pose transfer largely unexplored.

\begin{table}[t]
\centering
\small
\resizebox{\linewidth}{!}{
\begin{tabular}{l|c|c|c|c|c|c}
\hline
\textbf{Model} & \textbf{gt Cnd.} & \textbf{Text Cnd.} & \textbf{Exo Views} & \textbf{HOI Cnd.} & \textbf{3D Prior} & \textbf{Transitional} \\
\hline
\hline
Trj-Crafter \cite{yu2025trajectorycrafter}  & No  & Yes & 1 & No  & Yes & No \\
\hline
PMYS \cite{luo2024put}                & No  & No  & 1 & Yes & No & No \\
\hline
Exo2Ego-V \cite{liu2020exocentric}           & No  & No  & 4 & No  & Yes & No \\
\hline
EgoExo-Gen \cite{xu2025egoexo}              & Yes & Yes & 1 & Yes & No & No \\
\hline
Exo2EgoSyn \cite{mahdi2025exo2egosyn}              & No & No & 4 & No & No & No \\
\hline
\textbf{Ours}       & No  & No  & 1 & No  & No & Yes \\
\hline
\end{tabular}
}
\caption{Comparison of structural assumptions in different Exo2Ego methods. \textit{gt Cnd.} indicates use of any signal, e.g., the first frame, from ego ground-truth signals, \textit{Text Cnd.} indicates text-based conditioning, \textit{HOI Cnd.} denotes use of hand–object interaction information, and \textit{Transitional} shows if the model can generate exo-to-ego transitions.}
\label{tab:method_comparison}
\end{table}

\noindent{\textbf{Ego-Exo Cross-View Learning.}} Cross-view learning studies how models can relate, interpret, and synthesize observations of the same scene captured from different viewpoints, such as egocentric and exocentric perspectives. Exocentric understanding has been advanced by the emergence of ego--exo dual-view datasets - Ego-ExoLearn ~\cite{huang2024egoexolearn}, Ego-Exo4D~\cite{grauman2024ego}. Existing work has largely focused on perception and reasoning tasks, including ego--exo object correspondence~\cite{fu2025objectrelator, fu2025cross, mur2025mama, pan2025v2sam} and ego--exo visual question answering~\cite{he2025egoexobench, lee2025towards}. In this work, we instead address ego--exo video generation~\cite{liu2024exocentric}, which poses a substantially more challenging problem that requires synthesizing realistic egocentric observations rather than only understanding them.

Despite its importance, ego--exo video generation has received relatively limited attention. Luo et al.~\cite{luo2024put} propose a two-stage pipeline (PMYS) that first predicts hand trajectories and subsequently generates egocentric videos conditioned on exocentric inputs. EgoWorld~\cite{park2025egoworld} reconstructs egocentric views using exocentric depth, 3D hand pose estimates, and textual descriptions through point-cloud reprojection followed by diffusion-based inpainting. EgoExo-Gen~\cite{xu2025egoexo} conditions egocentric video synthesis on action descriptions together with the first egocentric frame. Complementary work studies the inverse direction of generation: intention-driven approaches focus on synthesizing exocentric videos from egocentric inputs~\cite{luo2024intention}. In our approach for exo-to-ego video synthesis, we only rely on camera poses, and do not require extra labels or building explicit representations, as demonstrated on Tab. \ref{tab:method_comparison}. Exo2Ego-V~\cite{liu2024exocentric} instead leverages multi-view exocentric observations and camera poses within a PixelNeRF-based diffusion framework to generate egocentric videos. In contrast, our method only requires a single exo view and no 3D prior, as shown on Tab. \ref{tab:method_comparison}.

Finally, Exo2EgoSyn~\cite{mahdi2025exo2egosyn} demonstrates how large pretrained video generators can be repurposed for the Exo2Ego task. While benefiting from strong pretrained representations, their approach relies on a single predicted egocentric frame to guide generation and applies per-frame camera control despite the model's temporally coupled attention. In contrast, we train a model better suited for per-frame pose conditioning and make use of full exocentric video inputs, while still leveraging large-scale pretrained knowledge through pose interpolation during data collection.

\begin{figure}[t]
  \centering
 \includegraphics[width=1.0\linewidth]{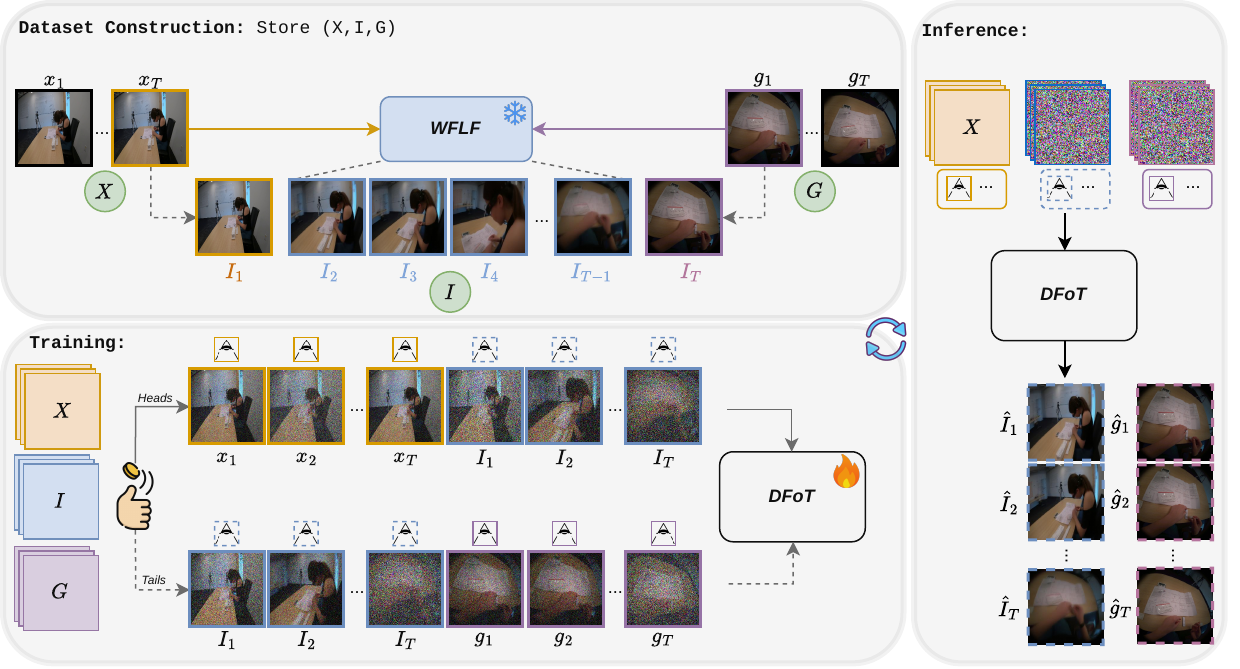}
  \caption{Syn2Seq-Forcing framework. Top-left: Using the last exo frame and first ego frame, we generate and cache pseudo-ground-truth interpolations with the WFLF model. Bottom-left: During training, a random pair of either (Exo–Interpolated) or (Interpolated–Ego) transitions is selected to train a DFoT conditioned on the corresponding camera poses. Right: During inference, the model receives the exo video and all camera poses to generate both the exo-to-ego transition and the final ego video.}
  \label{fig:framework}
  \vspace{-0.1in}
\end{figure}

\section{Methodology}
\subsection{Preliminaries}


\subsubsection{Diffusion Forcing Transformer.}

Diffusion Forcing Transformer (DFoT) is a video diffusion framework designed to support flexible history conditioning, \ie, the ability to condition generation on an arbitrary subset or length of previous frames. Unlike standard video diffusion pipelines that assume fixed conditioning layouts, DFoT applies the diffusion process with independent noise levels for each frame. This generalizes the \textit{noising-as-masking} principle: heavily noised frames effectively act as masked (removed) conditioning inputs, whereas lightly noised frames retain usable historical information, allowing the model to dynamically leverage past context.

This design enables a classifier-free guidance (CFG)-style formulation over the history during sampling. Specifically, the \textit{unconditional} branch corresponds to fully noising (masking) the historical frames, while the \textit{conditional} branch preserves selected history frames. Combining these score estimates provides a controllable trade-off between fidelity, temporal consistency, and diversity, now extended to handle variable-length and variable-structure histories. DFoT can be implemented using transformer-based video backbones (e.g., DiT or U-ViT) and is fully compatible with fine-tuning from pretrained video diffusion models, making it a flexible and practical choice for complex video generation tasks that require adaptive temporal conditioning.

Built on top of DFoT, History Guidance (HG) introduces structured mechanisms to exploit flexible history conditioning: \textit{Vanilla History Guidance (HG-v):} applies CFG with an arbitrary history window. \textit{Temporal History Guidance (HG-t):} combines score estimates from multiple history windows across time. \textit{Fractional History Guidance (HG-f):} uses partially noised history to modulate information by frequency content (e.g., low-pass or band-pass effects). \textit{HG-tf:} a joint time-frequency guidance strategy that combines HG-t and HG-f.

In Exo2Ego generation, we identify the primary challenge as the severe discontinuity between synchronized exocentric and egocentric sequences, both in visual content and camera poses. This abrupt spatio-temporal and geometric jump significantly degrades generation quality, as it violates the smooth-transition assumptions underlying standard video diffusion models. By explicitly bridging this gap through interpolation, we transform the synchronized pair into a single continuous signal. Such a formulation naturally aligns with DFoT, which is designed to flexibly process arbitrary history subsets within a unified sequence.

\subsubsection{WAN2.2 First/Last Frame (WFLF).}
WAN2.2 is a powerful foundational video Latent Diffusion Model(LDM) designed for high-quality video generation conditioned on text and/or images. Wan2.2-Lightning is a distilled Wan2.2 variant designed for efficient sampling with very few diffusion steps and without requiring CFG. WFLF is the boundary-conditioned baseline built on Wan2.2-Lightning 8-step LoRA, while following the FLF2V setup (conditioning on the first and last frames) to preserve global structure and temporal consistency. This design makes WFLF a compact and efficient frame interpolator, capable of generating temporally coherent transitions with minimal diffusion steps.
 
\subsection{Task Definition}
Given an exocentric video sequence $
X=\{x_1,\ldots,x_T\}\in\mathbb{R}^{T\times C\times H\times W},
$
with \(T\) frames, per-frame exocentric camera poses
$
P_{x}=\{p^{x}_1,\ldots,p^{x}_T\}\in\mathbb{R}^{T\times v},
$
and per-frame egocentric camera poses
$
P_{g}=\{p^{g}_1,\ldots,p^{g}_T\}\in\mathbb{R}^{T\times v},
$
our goal is to synthesize a temporally synchronized egocentric video
$
G=\{g_1,\ldots,g_T\}\in\mathbb{R}^{T\times C\times H\times W},
$
while preserving temporal alignment and scene-level consistency across viewpoints.

This cross-view generation problem is intrinsically challenging due to the large appearance and geometric gap between exocentric and egocentric observations, especially at the exo-to-ego transition (e.g., from \(x_T\) to \(g_1\)). The difficulty is further amplified by substantial viewpoint change and camera-motion mismatch between the two domains. In our setting, this becomes particularly severe because the exocentric camera is often static, i.e., \(p^{x}_1=\cdots=p^{x}_T\), which provides limited multi-view geometric cues and weak 3D observability. Consequently, the model must infer missing scene structure and plausible ego-view dynamics from highly under-constrained exocentric evidence.

\subsection{Our Framework}
Instead of learning a direct cross-view mapping
\[
G = f_\theta\!\left(X, P_x, P_g\right),
\]
we cast the task as \emph{diffusion-based sequential modeling} and propose \textit{Syn2Seq-Forcing}, as shown in Fig.~\ref{fig:framework}. We build a unified temporal stream \(S\) by concatenating exocentric frames, an intermediate interpolated segment, and egocentric frames. The interpolated segment is introduced to bridge the large exo-to-ego transition gap and enforce a smoother appearance and motion trajectory. In parallel, we interpolate camera poses to form a unified pose sequence
\[
P = (P_x, P_i, P_g),
\]
where \(P_i\) denotes interpolated transition poses. Following DFoT, we apply per-frame independent noise levels and feed the model the corrupted sequence \(S^n\), while conditioning on \(P\):
\[
\hat{\epsilon} = \mathrm{S2S}_{\phi}\!\left(S^n, P\right).
\]

This design offers three practical benefits. First, ego-view generation is conditioned not only on exo observations but also on transition-aware interpolated visual and pose cues, which is difficult to obtain in a direct one-shot formulation. Second, during sampling, the model jointly synthesizes both the transition segment and the target ego sequence in a temporally coherent manner. Third, because the formulation is sequence-centric rather than direction-specific, it naturally extends to the reverse setting (Ego\(\rightarrow\)Exo) within the same framework.

\subsubsection{Video Interpolation.}
To construct the unified stream \(S=[X, I, G]\), we synthesize an intermediate segment \(I\) that bridges the last exocentric frame \(x_T\) and the first egocentric frame \(g_1\). We use the frozen WFLF model for this purpose. Managing to require \(|I|=T\) (matching the lengths of \(X\) and \(G\)), we query WFLF to generate only the interior transition frames:
\[
I'=\mathrm{WFLF}(x_T, g_1,\text{prompt}), \qquad |I'|=T-2,
\]
and then explicitly attach the boundary frames:
\[
I=[x_T,\; I',\; g_1], \qquad |I|=T.
\]
Here, \(\textit{prompt}\) is a dataset-specific text prompt used to steer generation.

The resulting \(I\) is treated as a pseudo-interpolation signal: WFLF is not updated during training, and its outputs serve as supervisory transition targets within our S2S training pipeline. The details for querying WFLF is provided in supplementary materials.

\subsubsection{Pose Interpolation.}
Given \(P_x=\{p^x_t\}_{t=1}^{T}\) and \(P_g=\{p^g_t\}_{t=1}^{T}\), with \(p_t\in\mathbb{R}^{v}\), we define an intermediate pose trajectory
\[
P_i=\{p^i_j\}_{j=1}^{T},
\]
that connects the boundary poses \(p^x_T\) and \(p^g_1\). Each pose is decomposed as
\[
p=(k,\,[R\,|\,\mathbf{t}]),
\]
where \(k\) denotes camera intrinsics, \(R\in SO(3)\) is rotation, and \(\mathbf{t}\in\mathbb{R}^{3}\) is translation. For normalized time \(\tau_j=\frac{j-1}{T-1}\), we interpolate
\[
R^i_j=\mathrm{Slerp}\!\left(R^x_T,R^g_1;\tau_j\right),\qquad
\mathbf{t}^i_j=(1-\tau_j)\mathbf{t}^x_T+\tau_j\mathbf{t}^g_1,
\]
and form
\[
p^i_j=\big(k^i_j,\,[R^i_j\,|\,\mathbf{t}^i_j]\big),\quad j=1,\dots,T.
\]
In practice, \(k^i_j\) is fixed to the exocentric intrinsics for all \(j\). We also enforce boundary consistency:
\[
p^i_1=p^x_T,\qquad p^i_T=p^g_1.
\]
The final conditioning pose sequence is
\[
P=(P_x,\,P_i,\,P_g),
\]
which provides a smooth camera-geometry transition between the exocentric and egocentric segments. Details of \(\mathrm{Slerp}(\cdot)\) are provided in the supplementary material.

\subsubsection{Training.} We adopt a two-stage pretraining–finetuning paradigm, optimizing the following loss function for both stages:
\[
\mathcal{L} = \left\| \epsilon - \mathrm{S2S}_{\phi}\!\left(S^n, P\right) \right\|^2.
\]

\textit{Pretraining.} In the pretraining stage, the model is trained on 356k video clips collected from all available categories. Each sample consists of a direct Exo\(\rightarrow\)Ego pair, without any interpolation. The pretraining runs for 20 epochs (approximately 4 days) on 8 NVIDIA H200 GPUs, providing a strong initialization for subsequent fine-tuning.

\textit{Finetuning.} Building upon the pretrained weights, we finetune category-specific models using 40k videos per category, enabling interpolation, for 150 epochs (roughly 4 days) on 8 NVIDIA H200 GPUs. Training samples are constructed as paired triplets 
$(S=(X,I,G), P=(P_x,P_i,P_g)$
with \(I\) and \(P_i\) representing interpolated pseudo ground-truth visual and pose sequences. At each training step, we randomly select one of two sequential sub-tasks: either 
\[
D=\big((X,I),(P_x,P_i)\big), \quad \text{or} \quad D=\big((I,G),(P_i,P_g)\big).
\] 
This stochastic decomposition allows the model to jointly learn both transitions, Exo\(\rightarrow\)Interp and Interp\(\rightarrow\)Ego, within a unified training framework. More details regarding implementation, hyperparameters, and the setup for enabling Ego2Exo generation are provided in the supplementary material.

\subsubsection{Inference.}
During inference, we condition the model on clean exocentric frames and append a noise-only suffix of length \(2T\). The first \(T\) noise slots are denoised into the transition segment, and the remaining \(T\) noise slots are denoised into the egocentric segment. Formally, the generated sequence is
\[
\hat{S}=\big[X,\ \hat{I},\ \hat{G}\big],\qquad |\hat{I}|=|\hat{G}|=T,
\]
where \(\hat{I}\) and \(\hat{G}\) are produced in a single denoising process conditioned on \(X\) and the corresponding pose stream.

\section{Experiments}
\label{exps}

\subsection{Experimental Setup}
\noindent\textbf{Dataset.}
We evaluate our approach in three categories of the Ego-Exo4D~\cite{grauman2024ego} benchmark: Bike, Health, and Cooking. For each egocentric video, the dataset provides four exocentric videos (this study uses only one exo video randomly taken from the available ones) along with the corresponding camera poses for the fixed exocentric cameras and for every frame of the egocentric video.
Our experiments process 363 videos from the Bike category, 678 videos from the Cooking category, and 397 videos from the Health category. Both egocentric and exocentric videos are spatially resized to a resolution of \(H = W = 256\). Since the WFLF architecture requires the number of frames to satisfy \(4n + 1\), we sample 9 frames per video. We use a skip-frame step of 4 during preprocessing, ensuring that sufficient motion is captured in the sampled frames. For testing splits, we follow the partitions released by \cite{grauman2024ego}. 

\noindent\textbf{Baseline Construction.}\label{xxx}
To benchmark against prior Exo-to-Ego methods and reflect recent progress in controllable video generation, we include comparisons with Trajectory Crafter~\cite{yu2025trajectorycrafter}, Wan Fun Control~\cite{videoxfun2024}, and Wan VACE~\cite{jiang2025vace}, which utilize different conditioning strategies. Exo2Ego-V~\cite{liu2024exocentric} is limited to scenarios with four exocentric views, and EgoExo-Gen (X-Gen)~\cite{xu2025egoexo} lacks a publicly available implementation; therefore, both methods are omitted from our experiments.

\noindent\textbf{Evaluation Metrics.} To quantitatively assess our method, we report performance using PSNR, SSIM, and LPIPS (AlexNet~\cite{krizhevsky2012imagenet}), following the evaluation protocol described in \cite{liu2024exocentric}. This ensures a consistent and fair comparison with baseline methods while providing complementary measures of both pixel-level fidelity and perceptual similarity for the generated egocentric videos.

\subsection{Comparison with Baselines}
We present a comparison with the proposed baselines in Tab.~\ref{tab:main_table}. Across all categories, Syn2Seq-Forcing(ours) consistently achieves superior results, highlighting the robustness of our approach. To ensure a fair comparison with methods that cannot produce smooth exocentric-to-egocentric transitions, only the Ego-generated portion of Syn2Seq-Forcing is taken in to account for metric computation, excluding the interpolated segments. Qualitative results are shown in Fig.~\ref{fig:main_vis}, and additional visual examples are available in the supplementary material.

\begin{table*}[t]
\centering
\small
\begin{tabular}{l|ccc|ccc|ccc}
\toprule
\textbf{Method} 
& \multicolumn{3}{c|}{\textbf{Health}} 
& \multicolumn{3}{c|}{\textbf{Bike}} 
& \multicolumn{3}{c}{\textbf{Cooking}} \\
\cline{2-10}
& PSNR $\uparrow$ & SSIM $\uparrow$ & LPIPS $\downarrow$ 
& PSNR & SSIM & LPIPS 
& PSNR & SSIM & LPIPS \\
\hline
\hline
Trj-Crafter\cite{yu2025trajectorycrafter}
& 14.3091 & 0.4176 & 0.5623 
& 14.0102 & 0.3409 & 0.6001 
& 13.7554 & 0.3729 & 0.6003 \\
\hline
Wan-FCtrl\cite{videoxfun2024}
& 13.8993 & 0.4319 & 0.5734 
& 13.5699 & 0.3312 & 0.6197 
& 13.4200 & 0.3599 & 0.6071 \\
\hline
Wan VACE\cite{jiang2025vace}
& 14.1922 & 0.4299 & 0.5966 
& 13.2104 & 0.3329 & 0.6156 
& 13.3680 & 0.3689 & 0.6129 \\
\hline
Exo2EgoSyn\cite{mahdi2025exo2egosyn}
& 15.6234 & 0.4821 & 0.4991 
& 15.1319 & 0.3895 & 0.5052 
& 13.9011 & 0.4039 & 0.6125 \\
\hline
\textbf{Ours} 
& {16.7139} & {0.5728} & {0.4828} 
& {15.6301} & {0.4721} & {0.5012} 
& {14.3897} & {0.4531} & {0.5854} \\
\bottomrule
\end{tabular}
\caption{Main comparisons: Quantitative evaluation of Exo2Ego generation methods across three categories.}
\label{tab:main_table}
\end{table*}

\begin{figure}[p]
  \centering
 \includegraphics[width=1.0\linewidth]{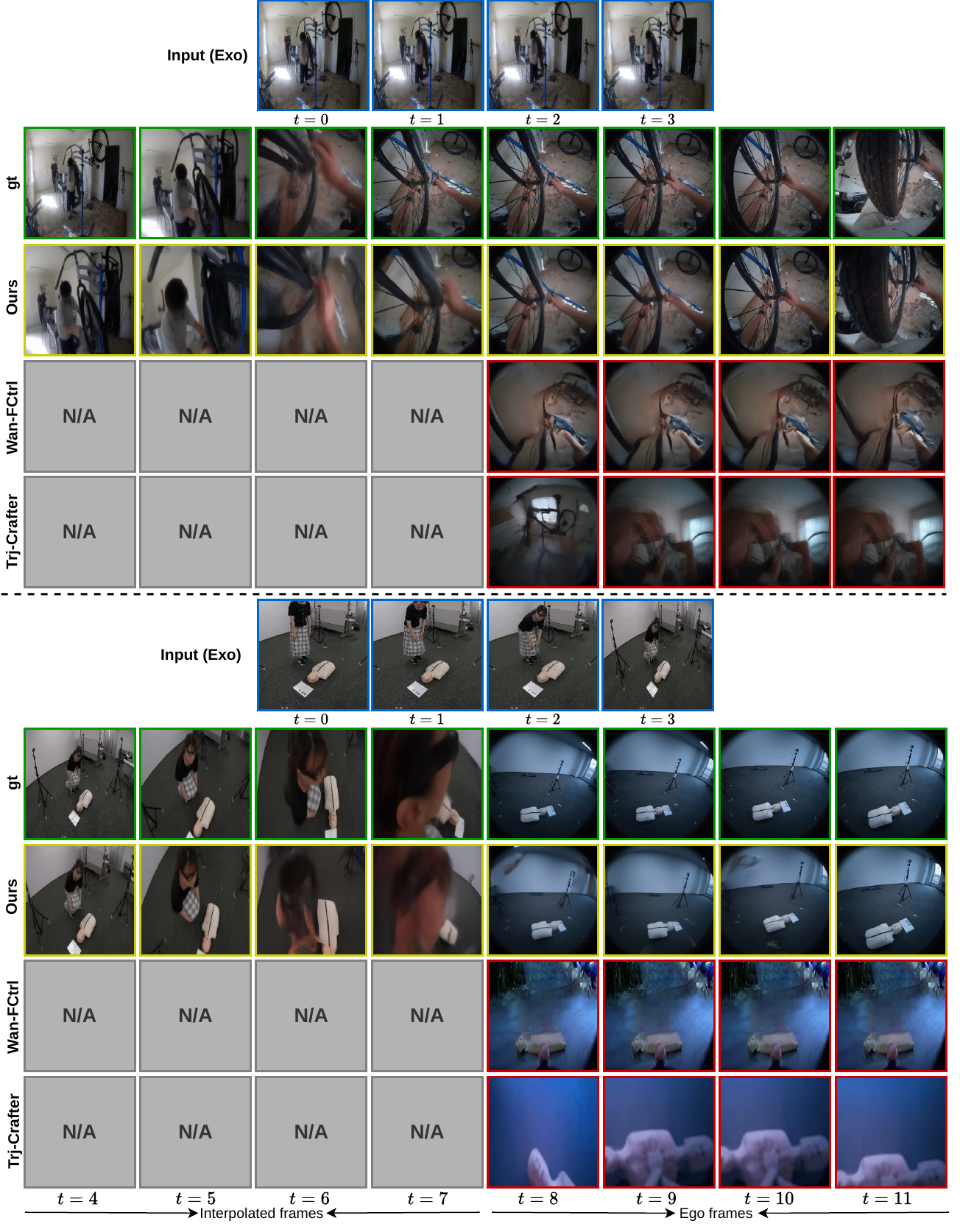}
  \caption{Comparison of different methods against ground-truth. Bettter viewed zoomed.}
  \label{fig:main_vis}
  \vspace{-0.1in}
\end{figure}

\subsection{Ablation Studies}
\subsubsection{Interpolation.} 
\begin{figure}[h]
  \centering
 \includegraphics[width=1.0\linewidth]{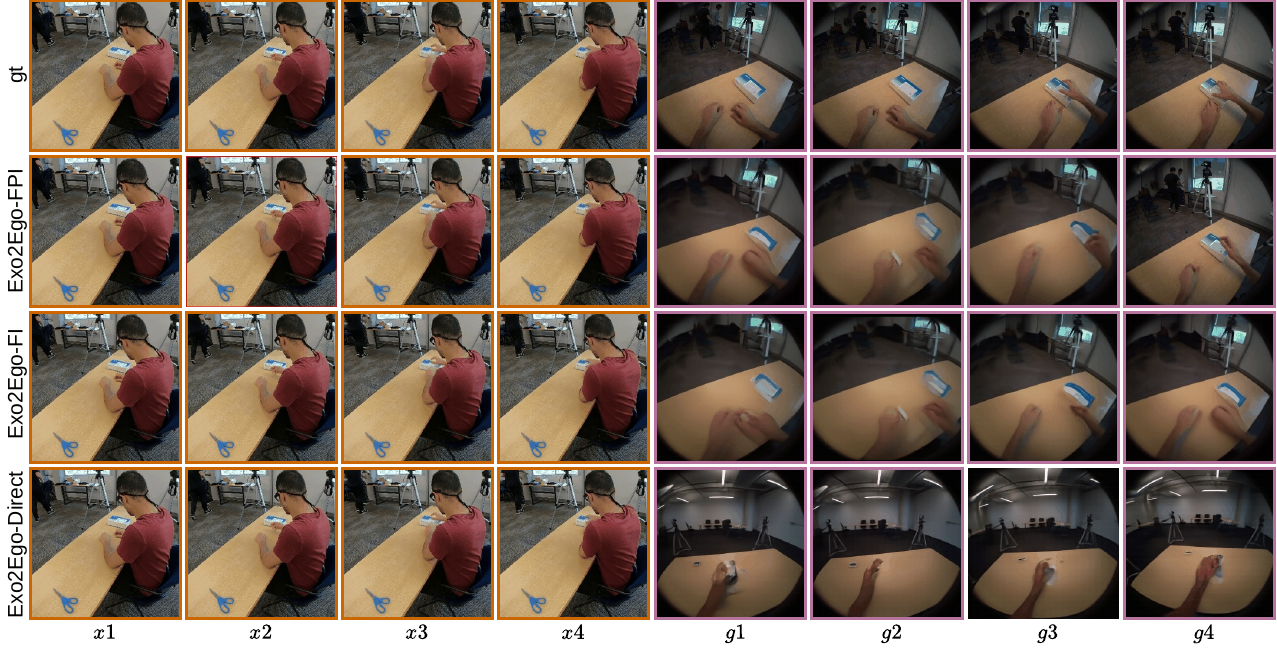}
  \caption{Comparison illustrating the effect of interpolation on the generated outputs.}
  \label{fig:abl_vis}
  \vspace{-0.1in}
\end{figure}

We demonstrate that interpolating only the video frames already provides a substantial performance boost. In this setup, no pose interpolation is performed; we use the egocentric poses directly and fill the remaining camera poses with zero vectors. We further evaluate the effect of including pose interpolation, with results summarized in Tab.~\ref{tab:intrp}. Qualitative examples illustrating the impact of both frame-only and frame-plus-pose interpolation are presented in Fig.~\ref{fig:abl_vis}.

\begin{table}[h!]
\centering
\begin{tabular}{l|ccc|ccc|ccc}
\toprule
\textbf{Model} & \multicolumn{3}{c|}{\textbf{Health}} & \multicolumn{3}{c|}{\textbf{Bike}} & \multicolumn{3}{c}{\textbf{Cooking}} \\
\cline{2-10}
 & PSNR $\uparrow$ & SSIM $\uparrow$ & LPIPS $\downarrow$ & PSNR & SSIM & LPIPS & PSNR & SSIM & LPIPS  \\
\hline
\hline
Exo2Ego-Direct & 13.5419 & 0.4176 & 0.5822 & 14.0091 & 0.3552 & 0.5866 & 13.1609 & 0.3831 & 0.6176 \\
\hline
Exo2Ego-FI & 16.1003 & 0.5427 & 0.4902 & 15.1245 & 0.4405 & 0.5387 & 14.2131 & 0.4401 & 0.5899 \\
\hline
Exo2Ego-FPI & 16.7139 & 0.5728 & 0.4828 & 15.6301 & 0.4721 & 0.5012 & 14.3897 & 0.4531 & 0.5854 \\
\bottomrule
\end{tabular}
\caption{Interpolation effect: \textit{Direct} predicts the ego view directly from the exo view without interpolation. \textit{FI} applies only frame interpolation, and uses ego camera poses with setting other poses to zero. \textit{FPI} applies both frame and pose interpolation.}
\label{tab:intrp}
\end{table}

\subsubsection{Type of Video Interpolation.} We, as reported in Tab.~\ref{tab:vid_int}, investigate the effectiveness of our interpolator, WFLF, in comparison to DFoT’s native inference-time interpolation. As mentioned earlier, typical video generator models alone cannot adequately handle the large viewpoint gap between exocentric and egocentric frames, resulting in inferior performance. We report quantitative results separately for the first $T$ generated frames (interpolated frames), the next $T$ frames (egocentric frames), and for both segments combined, providing a comprehensive evaluation of the interpolator’s contribution.

\begin{table}[h!]
\centering
\begin{tabular}{l|ccc|ccc|ccc}
\toprule

\textbf{Setup} & \multicolumn{3}{c|}{\textbf{Bike-INT}} & \multicolumn{3}{c|}{\textbf{Bike-EGO}} & \multicolumn{3}{c}{\textbf{Bike-Both}} \\
\cline{2-10}
 & PSNR $\uparrow$ & SSIM $\uparrow$ & LPIPS $\downarrow$ & PSNR & SSIM & LPIPS & PSNR & SSIM & LPIPS \\
\hline
\hline
WFLF & 15.6933 & 0.4754 & 0.5022 & 15.6301 & 0.4721 & 0.5012 & 15.6522 & 0.4736 & 0.5016 \\
\hline
DFoT-INTRPL & 13.1109 & 0.3203 & 0.6327 & 13.9012 & 0.3335 & 0.6192 & 13.4801 & 0.3276 & 0.6258 \\
\bottomrule
\end{tabular}
\caption{Video interpolator: DFoT-INTRPL denotes DFoT’s native interpolation capability during inference.}
\label{tab:vid_int}
\end{table}

\subsubsection{Type of Pose Embedding}
We perform an ablation study on the type of pose embedding, comparing three variants, according to Tab.~\ref{tab:pose_emb}: the \textit{Global} version, which represents camera poses as 16-dimensional vectors per frame; \textit{Ray Encoding}, which encodes per-pixel rays in a 180-dimensional space; and \textit{Plücker} embeddings, a per-pixel 6-dimensional representation of camera poses. Across experiments, we consistently observe, that Plücker embeddings yield the best performance, demonstrating superior compatibility with our interpolation framework.
\begin{table}[h!]
\centering
\begin{tabular}{l|ccc|ccc}
\toprule
\textbf{Setup} & \multicolumn{3}{c|}{\textbf{Bike-INT}} & \multicolumn{3}{c}{\textbf{Bike-EGO}} \\
\cline{2-7}
 & PSNR $\uparrow$ & SSIM $\uparrow$ & LPIPS $\downarrow$ & PSNR & SSIM & LPIPS \\
\hline
\hline
Global & 15.0799 & 0.4588 & 0.5116 & 15.0197 & 0.4571 & 0.5184 \\
\hline
Ray Encoding & 15.6028 & 0.4739 & 0.5097 & 15.5702 & 0.4729 & 0.5051 \\
\hline
Pluckers(ours) & 15.6933 & 0.4754 & 0.5022 & 15.6301 & 0.4721 & 0.5012 \\
\bottomrule
\end{tabular}
\caption{Effect of different methods for embedding camera poses.}
\label{tab:pose_emb}
\end{table}

\section{Conclusion}
\label{sec:conclusion}
In this work, we revisited Exo2Ego video generation task from a synchronization-aware perspective and demonstrated that the primary bottleneck is not simply the quality of conditioning, but the abrupt spatio-temporal and geometric discontinuities between synchronized exocentric and egocentric segments. To address this challenge, we replaced the traditional condition–output formulation with a sequential signal approach that explicitly bridges these gaps through interpolation and models the resulting sequence as a continuous signal under diffusion-based forcing. Exo2Ego generation typically depends on explicit depth reasoning, often implemented through 3D priors in prior work. In this work, we instead use interpolation as a lightweight proxy, guiding the model toward the target view via intermediate frames rather than explicit geometric estimation. This reformulation yields clear empirical improvements and remains effective even when only video interpolation is applied, highlighting that mitigating temporal discontinuity is a key factor driving performance.

Beyond these results, our framework provides a unified perspective on cross-view video generation: by treating the source, transitional, and target segments as a single structured sequence, it inherently supports flexible conditioning and points toward bidirectional view translation (Exo2Ego and Ego2Exo) within a single model. We anticipate that this approach will inspire future research on longer-horizon cross-view video synthesis and more comprehensive multi-view sequential modeling.

\section{Acknowledgements}
This research was partially funded by the Ministry of Education and Science of Bulgaria (support for INSAIT, part of
the Bulgarian National Roadmap for Research Infrastructure).
\bibliographystyle{splncs04}
\bibliography{main}

@String(ICCV  = {Int. Conf. Comput. Vis.})

@String(ICASSP=	{ICASSP})

@String(ICCV  = {ICCV})

@inproceedings{grauman2024ego,
  title={Ego-exo4d: Understanding skilled human activity from first-and third-person perspectives},
  author={Grauman, Kristen and Westbury, Andrew and Torresani, Lorenzo and Kitani, Kris and Malik, Jitendra and Afouras, Triantafyllos and Ashutosh, Kumar and Baiyya, Vijay and Bansal, Siddhant and Boote, Bikram and others},
  booktitle={Proceedings of the IEEE/CVF Conference on Computer Vision and Pattern Recognition},
  pages={19383--19400},
  year={2024}
}

@article{huang2025vipe,
  title={Vipe: Video pose engine for 3d geometric perception},
  author={Huang, Jiahui and Zhou, Qunjie and Rabeti, Hesam and Korovko, Aleksandr and Ling, Huan and Ren, Xuanchi and Shen, Tianchang and Gao, Jun and Slepichev, Dmitry and Lin, Chen-Hsuan and others},
  journal={arXiv preprint arXiv:2508.10934},
  year={2025}
}

@inproceedings{rombach2022high,
  title={High-resolution image synthesis with latent diffusion models},
  author={Rombach, Robin and Blattmann, Andreas and Lorenz, Dominik and Esser, Patrick and Ommer, Bj{\"o}rn},
  booktitle={Proceedings of the IEEE/CVF conference on computer vision and pattern recognition},
  pages={10684--10695},
  year={2022}
}

@article{pan2025v2sam,
  title={V²-SAM: Marrying SAM2 with Multi-Prompt Experts for Cross-View Object Correspondence},
  author={Pan, Jiancheng and Wang, Runze and Qian, Tianwen and Mahdi, Mohammad and Fu, Yanwei and Xue, Xiangyang and Huang, Xiaomeng and Van Gool, Luc and Paudel, Danda Pani and Fu, Yuqian},
  journal={arXiv preprint arXiv:2506.05856},
  year={2025}
}

@article{liu2025control,
  title={Control Copy-Paste: Controllable Diffusion-Based Augmentation Method for Remote Sensing Few-Shot Object Detection},
  author={Liu, Yanxing and Pan, Jiancheng and Zhang, Bingchen},
  journal={arXiv preprint arXiv:2507.21816},
  year={2025}
}

@article{liu2025diverse,
  title={Diverse Instance Generation via Diffusion Models for Enhanced Few-Shot Object Detection in Remote Sensing Images},
  author={Liu, Yanxing and Pan, Jiancheng and Yang, Jianwei and Chen, Tiancheng and Zhou, Peiling and Zhang, Bingchen},
  journal={IEEE Geoscience and Remote Sensing Letters},
  year={2025},
  publisher={IEEE}
}

@article{pan2025earthsynth,
  title={EarthSynth: Generating Informative Earth Observation with Diffusion Models},
  author={Pan, Jiancheng and Lei, Shiye and Fu, Yuqian and Li, Jiahao and Liu, Yanxing and Sun, Yuze and He, Xiao and Peng, Long and Huang, Xiaomeng and Zhao, Bo},
  journal={arXiv preprint arXiv:2505.12108},
  year={2025}
}

@inproceedings{huang2024egoexolearn,
  title={Egoexolearn: A dataset for bridging asynchronous ego-and exo-centric view of procedural activities in real world},
  author={Huang, Yifei and Chen, Guo and Xu, Jilan and Zhang, Mingfang and Yang, Lijin and Pei, Baoqi and Zhang, Hongjie and Dong, Lu and Wang, Yali and Wang, Limin and others},
  booktitle={Proceedings of the IEEE/CVF Conference on Computer Vision and Pattern Recognition},
  pages={22072--22086},
  year={2024}
}

@inproceedings{li2024egoexo,
  title={Egoexo-fitness: Towards egocentric and exocentric full-body action understanding},
  author={Li, Yuan-Ming and Huang, Wei-Jin and Wang, An-Lan and Zeng, Ling-An and Meng, Jing-Ke and Zheng, Wei-Shi},
  booktitle={European Conference on Computer Vision},
  pages={363--382},
  year={2024},
  organization={Springer}
}

@inproceedings{fu2025objectrelator,
  title={ObjectRelator: Enabling Cross-View Object Relation Understanding Across Ego-Centric and Exo-Centric Perspectives},
  author={Fu, Yuqian and Wang, Runze and Ren, Bin and Sun, Guolei and Gong, Biao and Fu, Yanwei and Paudel, Danda Pani and Huang, Xuanjing and Van Gool, Luc},
  booktitle={Proceedings of the IEEE/CVF International Conference on Computer Vision},
  pages={6530--6540},
  year={2025}
}

@inproceedings{liu2020exocentric,
  title={Exocentric to egocentric image generation via parallel generative adversarial network},
  author={Liu, Gaowen and Tang, Hao and Latapie, Hugo and Yan, Yan},
  booktitle={ICASSP 2020-2020 IEEE International Conference on Acoustics, Speech and Signal Processing (ICASSP)},
  pages={1843--1847},
  year={2020},
  organization={IEEE}
}

@article{park2025egoworld,
  title={EgoWorld: Translating Exocentric View to Egocentric View using Rich Exocentric Observations},
  author={Park, Junho and Ye, Andrew Sangwoo and Kwon, Taein},
  journal={arXiv preprint arXiv:2506.17896},
  year={2025}
}

@inproceedings{luo2024put,
  title={Put myself in your shoes: Lifting the egocentric perspective from exocentric videos},
  author={Luo, Mi and Xue, Zihui and Dimakis, Alex and Grauman, Kristen},
  booktitle={European Conference on Computer Vision},
  pages={407--425},
  year={2024},
  organization={Springer}
}

@article{xu2025egoexo,
  title={Egoexo-gen: Ego-centric video prediction by watching exo-centric videos},
  author={Xu, Jilan and Huang, Yifei and Pei, Baoqi and Hou, Junlin and Li, Qingqiu and Chen, Guo and Zhang, Yuejie and Feng, Rui and Xie, Weidi},
  journal={arXiv preprint arXiv:2504.11732},
  year={2025}
}

@article{luo2024intention,
  title={Intention-driven ego-to-exo video generation},
  author={Luo, Hongchen and Zhu, Kai and Zhai, Wei and Cao, Yang},
  journal={arXiv preprint arXiv:2403.09194},
  year={2024}
}

@article{liu2024exocentric,
  title={Exocentric-to-egocentric video generation},
  author={Liu, Jia-Wei and Mao, Weijia and Xu, Zhongcong and Keppo, Jussi and Shou, Mike Zheng},
  journal={Advances in Neural Information Processing Systems},
  volume={37},
  pages={136149--136172},
  year={2024}
}

@inproceedings{jiang2025vace,
  title={Vace: All-in-one video creation and editing},
  author={Jiang, Zeyinzi and Han, Zhen and Mao, Chaojie and Zhang, Jingfeng and Pan, Yulin and Liu, Yu},
  booktitle={Proceedings of the IEEE/CVF International Conference on Computer Vision},
  pages={17191--17202},
  year={2025}
}

@misc{videoxfun2024,
  author       = {{AIGC-Apps}},
  title        = {VideoX-Fun: A Flexible Framework for Video Generation},
  year         = {2024},
  howpublished = {\url{https://github.com/aigc-apps/VideoX-Fun}},
  note         = {Accessed: 2026-03-05}
}

@inproceedings{blattmann2023align,
  title={Align your latents: High-resolution video synthesis with latent diffusion models},
  author={Blattmann, Andreas and Rombach, Robin and Ling, Huan and Dockhorn, Tim and Kim, Seung Wook and Fidler, Sanja and Kreis, Karsten},
  booktitle={Proceedings of the IEEE/CVF conference on computer vision and pattern recognition},
  pages={22563--22575},
  year={2023}
}

@article{guo2023animatediff,
  title={Animatediff: Animate your personalized text-to-image diffusion models without specific tuning},
  author={Guo, Yuwei and Yang, Ceyuan and Rao, Anyi and Liang, Zhengyang and Wang, Yaohui and Qiao, Yu and Agrawala, Maneesh and Lin, Dahua and Dai, Bo},
  journal={arXiv preprint arXiv:2307.04725},
  year={2023}
}

@article{ho2022imagen,
  title={Imagen video: High definition video generation with diffusion models},
  author={Ho, Jonathan and Chan, William and Saharia, Chitwan and Whang, Jay and Gao, Ruiqi and Gritsenko, Alexey and Kingma, Diederik P and Poole, Ben and Norouzi, Mohammad and Fleet, David J and others},
  journal={arXiv preprint arXiv:2210.02303},
  year={2022}
}

@article{ho2022video,
  title={Video diffusion models},
  author={Ho, Jonathan and Salimans, Tim and Gritsenko, Alexey and Chan, William and Norouzi, Mohammad and Fleet, David J},
  journal={Advances in neural information processing systems},
  volume={35},
  pages={8633--8646},
  year={2022}
}

@inproceedings{wu2023tune,
  title={Tune-a-video: One-shot tuning of image diffusion models for text-to-video generation},
  author={Wu, Jay Zhangjie and Ge, Yixiao and Wang, Xintao and Lei, Stan Weixian and Gu, Yuchao and Shi, Yufei and Hsu, Wynne and Shan, Ying and Qie, Xiaohu and Shou, Mike Zheng},
  booktitle={Proceedings of the IEEE/CVF international conference on computer vision},
  pages={7623--7633},
  year={2023}
}

@article{zhang2025show,
  title={Show-1: Marrying pixel and latent diffusion models for text-to-video generation},
  author={Zhang, David Junhao and Wu, Jay Zhangjie and Liu, Jia-Wei and Zhao, Rui and Ran, Lingmin and Gu, Yuchao and Gao, Difei and Shou, Mike Zheng},
  journal={International Journal of Computer Vision},
  volume={133},
  number={4},
  pages={1879--1893},
  year={2025},
  publisher={Springer}
}

@article{zhou2022magicvideo,
  title={Magicvideo: Efficient video generation with latent diffusion models},
  author={Zhou, Daquan and Wang, Weimin and Yan, Hanshu and Lv, Weiwei and Zhu, Yizhe and Feng, Jiashi},
  journal={arXiv preprint arXiv:2211.11018},
  year={2022}
}

@article{wan2025wan,
  title={Wan: Open and advanced large-scale video generative models},
  author={Wan, Team and Wang, Ang and Ai, Baole and Wen, Bin and Mao, Chaojie and Xie, Chen-Wei and Chen, Di and Yu, Feiwu and Zhao, Haiming and Yang, Jianxiao and others},
  journal={arXiv preprint arXiv:2503.20314},
  year={2025}
}

@article{lee2025towards,
  title={Towards Comprehensive Scene Understanding: Integrating First and Third-Person Views for LVLMs},
  author={Lee, Insu and Park, Wooje and Jang, Jaeyun and Noh, Minyoung and Shim, Kyuhong and Shim, Byonghyo},
  journal={arXiv preprint arXiv:2505.21955},
  year={2025}
}

@article{he2025egoexobench,
  title={Egoexobench: A benchmark for first-and third-person view video understanding in mllms},
  author={He, Yuping and Huang, Yifei and Chen, Guo and Pei, Baoqi and Xu, Jilan and Lu, Tong and Pang, Jiangmiao},
  journal={arXiv preprint arXiv:2507.18342},
  year={2025}
}

@misc{fu2025cross,
  title={Cross-View Multi-Modal Segmentation @ Ego-Exo4D Challenges 2025},
  author={Fu, Yuqian and Wang, Runze and Fu, Yanwei and Paudel, Danda Pani and Van Gool, Luc},
  year={2025},
  note={Ego-Exo4D Challenge}
}

@article{blattmann2023stable,
  title={Stable video diffusion: Scaling latent video diffusion models to large datasets},
  author={Blattmann, Andreas and Dockhorn, Tim and Kulal, Sumith and Mendelevitch, Daniel and Kilian, Maciej and Lorenz, Dominik and Levi, Yam and English, Zion and Voleti, Vikram and Letts, Adam and others},
  journal={arXiv preprint arXiv:2311.15127},
  year={2023}
}

@inproceedings{mur2025mama,
  title={O-MaMa: Learning Object Mask Matching between Egocentric and Exocentric Views},
  author={Mur-Labadia, Lorenzo and Santos-Villafranca, Maria and Bermudez-Cameo, Jesus and Perez-Yus, Alejandro and Martinez-Cantin, Ruben and Guerrero, Jose J},
  booktitle={ICCV},
  year={2025}
}

@article{krizhevsky2012imagenet,
  title={Imagenet classification with deep convolutional neural networks},
  author={Krizhevsky, Alex and Sutskever, Ilya and Hinton, Geoffrey E},
  journal={Advances in neural information processing systems},
  volume={25},
  year={2012}
}

@article{yang2024cogvideox,
  title={Cogvideox: Text-to-video diffusion models with an expert transformer},
  author={Yang, Zhuoyi and Teng, Jiayan and Zheng, Wendi and Ding, Ming and Huang, Shiyu and Xu, Jiazheng and Yang, Yuanming and Hong, Wenyi and Zhang, Xiaohan and Feng, Guanyu and others},
  journal={arXiv preprint arXiv:2408.06072},
  year={2024}
}

@inproceedings{peebles2023scalable,
  title={Scalable diffusion models with transformers},
  author={Peebles, William and Xie, Saining},
  booktitle={Proceedings of the IEEE/CVF international conference on computer vision},
  pages={4195--4205},
  year={2023}
}

@inproceedings{huang2024learning,
  title={Learning disentangled identifiers for action-customized text-to-image generation},
  author={Huang, Siteng and Gong, Biao and Feng, Yutong and Chen, Xi and Fu, Yuqian and Liu, Yu and Wang, Donglin},
  booktitle={Proceedings of the IEEE/CVF Conference on Computer Vision and Pattern Recognition},
  pages={7797--7806},
  year={2024}
}

@article{bai2025recammaster,
  title={Recammaster: Camera-controlled generative rendering from a single video},
  author={Bai, Jianhong and Xia, Menghan and Fu, Xiao and Wang, Xintao and Mu, Lianrui and Cao, Jinwen and Liu, Zuozhu and Hu, Haoji and Bai, Xiang and Wan, Pengfei and others},
  journal={arXiv preprint arXiv:2503.11647},
  year={2025}
}

@article{sun2025worldplay,
  title={Worldplay: Towards long-term geometric consistency for real-time interactive world modeling},
  author={Sun, Wenqiang and Zhang, Haiyu and Wang, Haoyuan and Wu, Junta and Wang, Zehan and Wang, Zhenwei and Wang, Yunhong and Zhang, Jun and Wang, Tengfei and Guo, Chunchao},
  journal={arXiv preprint arXiv:2512.14614},
  year={2025}
}

@article{he2024cameractrl,
  title={Cameractrl: Enabling camera control for text-to-video generation},
  author={He, Hao and Xu, Yinghao and Guo, Yuwei and Wetzstein, Gordon and Dai, Bo and Li, Hongsheng and Yang, Ceyuan},
  journal={arXiv preprint arXiv:2404.02101},
  year={2024}
}

@inproceedings{wang2024motionctrl,
  title={Motionctrl: A unified and flexible motion controller for video generation},
  author={Wang, Zhouxia and Yuan, Ziyang and Wang, Xintao and Li, Yaowei and Chen, Tianshui and Xia, Menghan and Luo, Ping and Shan, Ying},
  booktitle={ACM SIGGRAPH 2024 Conference Papers},
  pages={1--11},
  year={2024}
}

@inproceedings{yang2024direct,
  title={Direct-a-video: Customized video generation with user-directed camera movement and object motion},
  author={Yang, Shiyuan and Hou, Liang and Huang, Haibin and Ma, Chongyang and Wan, Pengfei and Zhang, Di and Chen, Xiaodong and Liao, Jing},
  booktitle={ACM SIGGRAPH 2024 Conference Papers},
  pages={1--12},
  year={2024}
}

@article{hou2024training,
  title={Training-free camera control for video generation},
  author={Hou, Chen and Chen, Zhibo},
  journal={arXiv preprint arXiv:2406.10126},
  year={2024}
}

@article{team2026advancing,
  title={Advancing Open-source World Models},
  author={Team, Robbyant and Gao, Zelin and Wang, Qiuyu and Zeng, Yanhong and Zhu, Jiapeng and Cheng, Ka Leong and Li, Yixuan and Wang, Hanlin and Xu, Yinghao and Ma, Shuailei and others},
  journal={arXiv preprint arXiv:2601.20540},
  year={2026}
}

@article{parker2025genie,
  title={Genie 3: A new frontier for world models},
  author={Parker-Holder, Jack and Fruchter, Shlomi},
  journal={URL https://deepmind. google/discover/blog/genie-3-a-new-frontier-for-world-models/. Blog post},
  year={2025}
}

@article{song2025history,
  title={History-guided video diffusion},
  author={Song, Kiwhan and Chen, Boyuan and Simchowitz, Max and Du, Yilun and Tedrake, Russ and Sitzmann, Vincent},
  journal={arXiv preprint arXiv:2502.06764},
  year={2025}
}

@article{huang2025self,
  title={Self forcing: Bridging the train-test gap in autoregressive video diffusion},
  author={Huang, Xun and Li, Zhengqi and He, Guande and Zhou, Mingyuan and Shechtman, Eli},
  journal={arXiv preprint arXiv:2506.08009},
  year={2025}
}

@article{cui2025self,
  title={Self-forcing++: Towards minute-scale high-quality video generation},
  author={Cui, Justin and Wu, Jie and Li, Ming and Yang, Tao and Li, Xiaojie and Wang, Rui and Bai, Andrew and Ban, Yuanhao and Hsieh, Cho-Jui},
  journal={arXiv preprint arXiv:2510.02283},
  year={2025}
}

@article{shin2025motionstream,
  title={Motionstream: Real-time video generation with interactive motion controls},
  author={Shin, Joonghyuk and Li, Zhengqi and Zhang, Richard and Zhu, Jun-Yan and Park, Jaesik and Shechtman, Eli and Huang, Xun},
  journal={arXiv preprint arXiv:2511.01266},
  year={2025}
}

@article{mahdi2025exo2egosyn,
  title={Exo2EgoSyn: Unlocking Foundation Video Generation Models for Exocentric-to-Egocentric Video Synthesis},
  author={Mahdi, Mohammad and Fu, Yuqian and Savov, Nedko and Pan, Jiancheng and Paudel, Danda Pani and Van Gool, Luc},
  journal={arXiv preprint arXiv:2511.20186},
  year={2025}
}

@article{hacohen2026ltx,
  title={LTX-2: Efficient Joint Audio-Visual Foundation Model},
  author={HaCohen, Yoav and Brazowski, Benny and Chiprut, Nisan and Bitterman, Yaki and Kvochko, Andrew and Berkowitz, Avishai and Shalem, Daniel and Lifschitz, Daphna and Moshe, Dudu and Porat, Eitan and others},
  journal={arXiv preprint arXiv:2601.03233},
  year={2026}
}

@article{kong2024hunyuanvideo,
  title={Hunyuanvideo: A systematic framework for large video generative models},
  author={Kong, Weijie and Tian, Qi and Zhang, Zijian and Min, Rox and Dai, Zuozhuo and Zhou, Jin and Xiong, Jiangfeng and Li, Xin and Wu, Bo and Zhang, Jianwei and others},
  journal={arXiv preprint arXiv:2412.03603},
  year={2024}
}

@inproceedings{yu2025trajectorycrafter,
  title={Trajectorycrafter: Redirecting camera trajectory for monocular videos via diffusion models},
  author={Yu, Mark and Hu, Wenbo and Xing, Jinbo and Shan, Ying},
  booktitle={Proceedings of the IEEE/CVF international conference on computer vision},
  pages={100--111},
  year={2025}
}

@article{wu2025video,
  title={Video world models with long-term spatial memory},
  author={Wu, Tong and Yang, Shuai and Po, Ryan and Xu, Yinghao and Liu, Ziwei and Lin, Dahua and Wetzstein, Gordon},
  journal={arXiv preprint arXiv:2506.05284},
  year={2025}
}

@inproceedings{ren2025gen3c,
  title={Gen3c: 3d-informed world-consistent video generation with precise camera control},
  author={Ren, Xuanchi and Shen, Tianchang and Huang, Jiahui and Ling, Huan and Lu, Yifan and Nimier-David, Merlin and M{\"u}ller, Thomas and Keller, Alexander and Fidler, Sanja and Gao, Jun},
  booktitle={Proceedings of the IEEE/CVF Conference on Computer Vision and Pattern Recognition},
  pages={6121--6132},
  year={2025}
}

@inproceedings{li2025vmem,
  title={Vmem: Consistent interactive video scene generation with surfel-indexed view memory},
  author={Li, Runjia and Torr, Philip and Vedaldi, Andrea and Jakab, Tomas},
  booktitle={Proceedings of the IEEE/CVF International Conference on Computer Vision},
  pages={25690--25699},
  year={2025}
}

\title{From Synchrony to Sequence: Exo-to-Ego Generation via Interpolation}

\titlerunning{Syn2Seq-Forcing}

\author{Mohammad Mahdi\inst{1}\thanks{\email{firstname.lastname@insait.ai}} \quad Nedko Savov\inst{1} \quad
Danda Pani Paudel\inst{1}  \quad Luc Van Gool\inst{1}}
\authorrunning{Mahdi et al.}

\institute{$^1$INSAIT, Sofia University “St. Kliment Ohridski” \\
\url{https://github.com/insait-institute/Syn2Seq}}
\authorrunning{Mahdi et al.}

\maketitle
\section{Frame Interpolator, WFLF}
WAN2.2~\cite{wan2025wan} is a powerful foundational video Latent Diffusion Model (LDM) designed for high-quality video generation conditioned on text and/or images. The model supports rich spatiotemporal generation and has demonstrated strong performance across diverse motion and scene dynamics. Wan2.2-Lightning is a distilled variant of the Wan2.2 family that is optimized for efficient inference. Through distillation, it significantly reduces the number of required diffusion steps while maintaining visual fidelity and motion quality. In particular, Wan2.2-Lightning enables fast video generation with only a small number of sampling steps and does not require classifier-free guidance (CFG), making it suitable for lightweight generation pipelines.

In our experiments, we adopt the \textit{WAN2.2 First/Last Frame} (WFLF) configuration, which follows the FLF2V setup where the model is conditioned on the first and last frames of a video sequence. The model then generates the intermediate frames that connect these two boundary conditions. We implement WFLF using the Wan2.2-Lightning 8-step LoRA variant within the Diffusers framework, which provides an efficient and stable inference pipeline.

\paragraph{Pseudo-Label Frame Generation with WFLF.}
To generate pseudo-labeled interpolated frames using the WFLF pipeline, we condition the model on the last frame of the exocentric (exo) video and the first frame of the egocentric (ego) video. These two boundary frames define the endpoints of the transition, and WFLF is tasked with generating the intermediate frames that smoothly connect them. Importantly, no camera pose information is provided during this process. Instead, the generation relies solely on the visual boundary conditions and a textual prompt describing the intended transition from third-person to first-person perspective.

Because the scenes in our experiments belong to distinct semantic categories, we use category-specific prompts to guide the generation. These prompts remain fixed for all samples within each category to ensure consistency across the generated pseudo-labels. Below we list the prompts used for each category.

\paragraph{Prompt for \textit{Bike} category.}
\begin{quote}
\textit{A man is fixing a bike. The camera starts in third-person view and smoothly moves toward the person, pushing in until it reaches his head and seamlessly transitions into a first-person perspective, showing the bike from his point of view.}
\end{quote}

\paragraph{Prompt for \textit{Cooking} category.}
\begin{quote}
\textit{A person is cooking. The camera starts in third-person view and smoothly moves toward the person, pushing in until it reaches his head and seamlessly transitions into a first-person perspective, showing the environment from his point of view.}
\end{quote}

\paragraph{Prompt for \textit{Health CPR} category.}
\begin{quote}
\textit{A person is performing Cardiopulmonary Resuscitation (CPR) on someone. The camera starts in third-person view and smoothly moves toward the person, pushing in until it reaches their head and seamlessly transitions into a first-person perspective, showing the CPR action from their point of view.}
\end{quote}

\paragraph{Prompt for \textit{Health Covid} category.}
\begin{quote}
\textit{A person is taking a COVID test. The camera starts in third-person view and smoothly moves toward the person, pushing in until it reaches their head and seamlessly transitions into a first-person perspective, showing the test kit and surroundings from their point of view.}
\end{quote}

\section{Rotation Interpolator: Spherical Linear Interpolation (SLERP)}

To obtain smooth transitions between camera orientations, we interpolate the rotations between the last exocentric frame and the first egocentric frame using \textit{Spherical Linear Interpolation (SLERP)}. SLERP is a standard technique for interpolating rotations represented as unit quaternions. Unlike naive linear interpolation in Euclidean space, SLERP respects the geometry of the unit hypersphere on which quaternions lie, producing constant angular velocity and smooth rotational motion between two orientations.

Let $q_0, q_1 \in \mathbb{R}^4$ denote two unit quaternions representing the starting and ending rotations, respectively. These quaternions lie on the unit 3-sphere $\mathbb{S}^3$. SLERP computes intermediate rotations by moving along the shortest geodesic path on this sphere. Given an interpolation parameter $t \in [0,1]$, the interpolated quaternion $q(t)$ is defined as

\begin{equation}
q(t) = 
\frac{\sin((1-t)\theta)}{\sin(\theta)} q_0 +
\frac{\sin(t\theta)}{\sin(\theta)} q_1,
\end{equation}

where $\theta$ is the angular distance between the two quaternions:

\begin{equation}
\theta = \arccos(\langle q_0, q_1 \rangle),
\end{equation}

and $\langle q_0, q_1 \rangle$ denotes the dot product between the quaternions.

Because unit quaternions $q$ and $-q$ represent the same rotation, we first ensure interpolation follows the shortest path on the sphere. If $\langle q_0, q_1 \rangle < 0$, we negate one quaternion (e.g., $q_1 \leftarrow -q_1$), which keeps the represented rotation unchanged but avoids traversing the long arc on $\mathbb{S}^3$.

For numerical stability, when the two quaternions are extremely close ($\langle q_0, q_1 \rangle \approx 1$), the denominator $\sin(\theta)$ becomes very small. In this case we fall back to normalized linear interpolation:

\begin{equation}
q(t) = \frac{(1-t)q_0 + t q_1}{\|(1-t)q_0 + t q_1\|}.
\end{equation}

In our pipeline, the rotation matrices from the boundary frames are first converted to unit quaternions. Given the quaternion of the last exocentric frame $q_{\text{exo}}$ and the quaternion of the first egocentric frame $q_{\text{ego}}$, we sample a set of interpolation coefficients $t_1,\dots,t_N$ and apply SLERP to generate a sequence of intermediate quaternions. These interpolated quaternions are then converted back to rotation matrices and combined with the corresponding translations to form the camera poses used in our interpolated sequence.

\section{Framework Setup}

Our framework consists of three main stages: \textit{(1) Dataset Construction}, \textit{(2) Training}, and \textit{(3) Inference}. In this section, we describe the practical setup and implementation details used for each stage of the pipeline.

\subsection{Dataset Construction}

For each of the three evaluated categories, we construct a dataset containing 40k samples. Each sample consists of three components: the source exocentric video, the interpolated transition frames generated by WFLF, and the target egocentric video. These samples form the training data used during the category-specific fine-tuning stage.

To obtain high-quality interpolations from WFLF, we found it beneficial to perform generation at a higher spatial resolution. Specifically, we generate frames at a resolution of $H=480$ and $W=832$. After generation, the interpolated frames are resized to the training resolution of $H=W=256$ to match the input resolution used by the video diffusion model.

For each sequence, we query the WFLF model to generate 49 frames between the boundary frames. From these generated frames, we uniformly sample 9 frames to construct the interpolated transition segment used in our training data.

The dataset generation process for each category takes approximately 1 day using 8 NVIDIA H200 GPUs.

\subsection{Training}

For the training stage, we largely follow the setup proposed in the Diffusion Forcing Transformer (DFoT)~\cite{song2025history} framework, adapting it to our Exo2Ego scenario. Our sequences are trained with a maximum token length of 18 frames, of which 9 frames are designated as context frames. This configuration allows the model to effectively leverage temporal dependencies while generating future frames.

We employ a batch size of 8 per GPU and distribute training across 8 GPUs, resulting in an effective batch size of 64. This setup ensures stable optimization and sufficient gradient statistics for learning across long video sequences.

For the model backbone, we use a 3D variant of the U-ViT architecture (U-ViT3D), which is capable of incorporating additional conditioning information—in our case, per-frame camera poses—using the Feature-wise Linear Modulation (FiLM) technique. The FiLM layers condition the hidden features on the camera pose signals, allowing the model to adapt its representation to geometric changes across views.

Within the intermediate blocks of the UNet backbone, we perform self-attention between all video token embeddings, enabling the model to capture long-range temporal and spatial dependencies. For the earliest UNet blocks, which operate on high-resolution, low-level features, we instead employ standard ResNet layers to efficiently process fine-grained spatial details before projecting them into the attention-based space. We optimize the model using the AdamW optimizer with a learning rate of $5\times10^{-5}$, carefully balancing convergence speed and stability.

For iterations where we train on the Interp→Ego sequence, we condition on VAE-encoded features of the Exo video as an additional input. In contrast, for Exo→Interp sequences, this condition is replaced with VAE-encoded features of black frames. We use the Wan2.1 VAE for all feature encoding.

\subsection{Inference}
To guide generation, we apply Fractional History Guidance (HG-f) with a
scale factor of 3.0. This guidance modulates the influence of previous frames across different frequency components, improving temporal coherence and preserving motion dynamics in the generated sequence. The diffusion process is run for 50 backward steps, providing sufficient denoising while maintaining fidelity to the context frames.

\subsection{Ego-to-Exo Generation}
Our current formulation for Exo2Ego generation operates on ordered sequences ($X_{1:N} \rightarrow I_{1:N} \rightarrow G_{1:N}$). For Ego2Exo training, the sequence can be reversed ($G_{N:1} \rightarrow I_{N:1} \rightarrow X_{N:1}$) and then flipped back at inference, making the framework naturally compatible with both directions without architectural modifications or additional pseudo-ground truths. We briefly explore this extension on the \textit{bike} subcategory and report qualitative results in the following section.

\section{Visual Examples}

\begin{figure}[h]
 \includegraphics[width=1.0\linewidth]{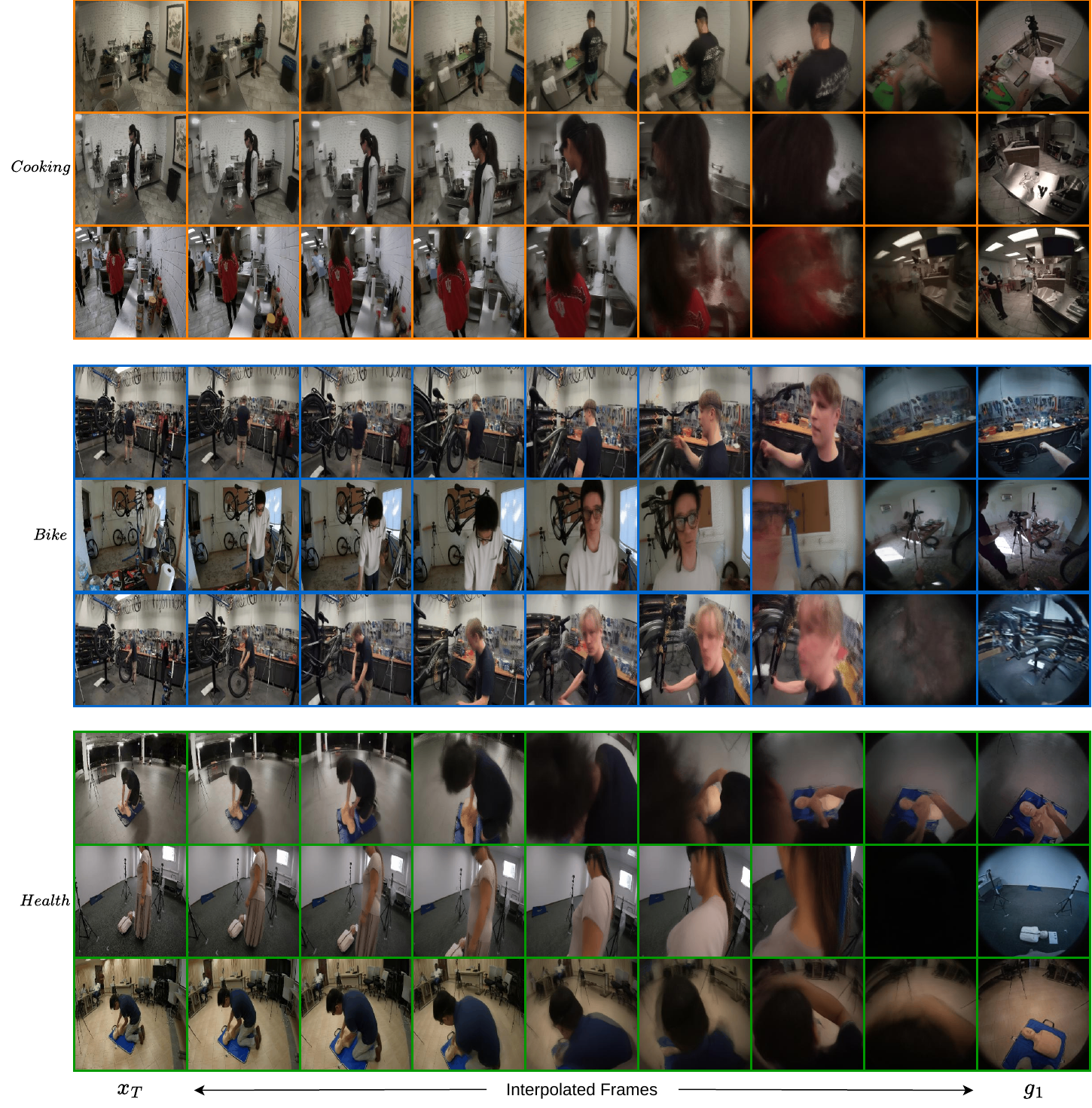}
  \caption{The model's ability to generate intermediate frames bridging the last frame of the exocentric video and the first frame of the egocentric video.}
  \label{fig:interpolation}
  \vspace{-0.1in}
\end{figure}

\begin{figure}[h]
 \includegraphics[width=1.0\linewidth]{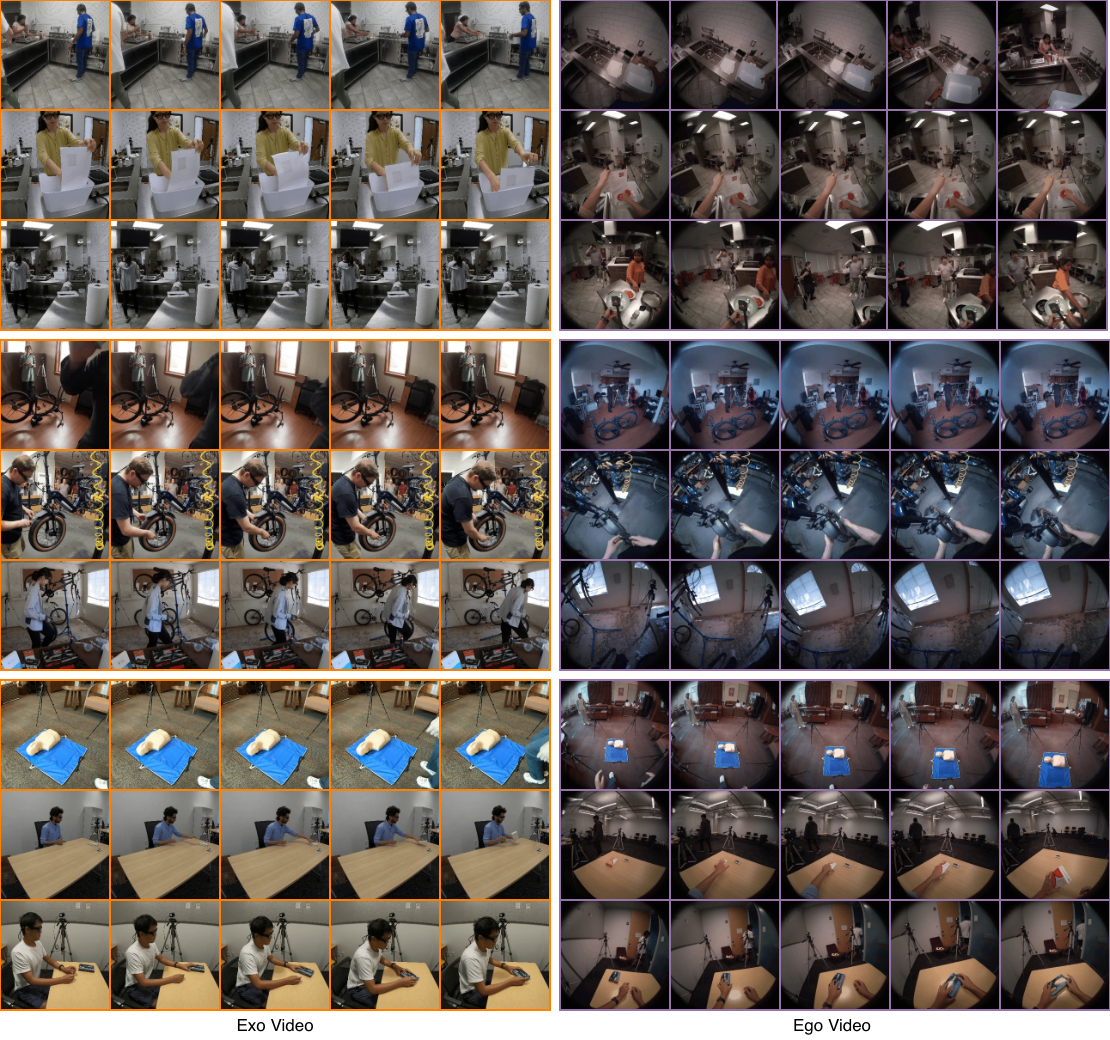}
  \caption{The model's ability to perform cross-view generation of egocentric video from exocentric video.}
  \label{fig:exo2ego}
  \vspace{-0.1in}
\end{figure}

\begin{figure}[h]
 \includegraphics[width=1.0\linewidth]{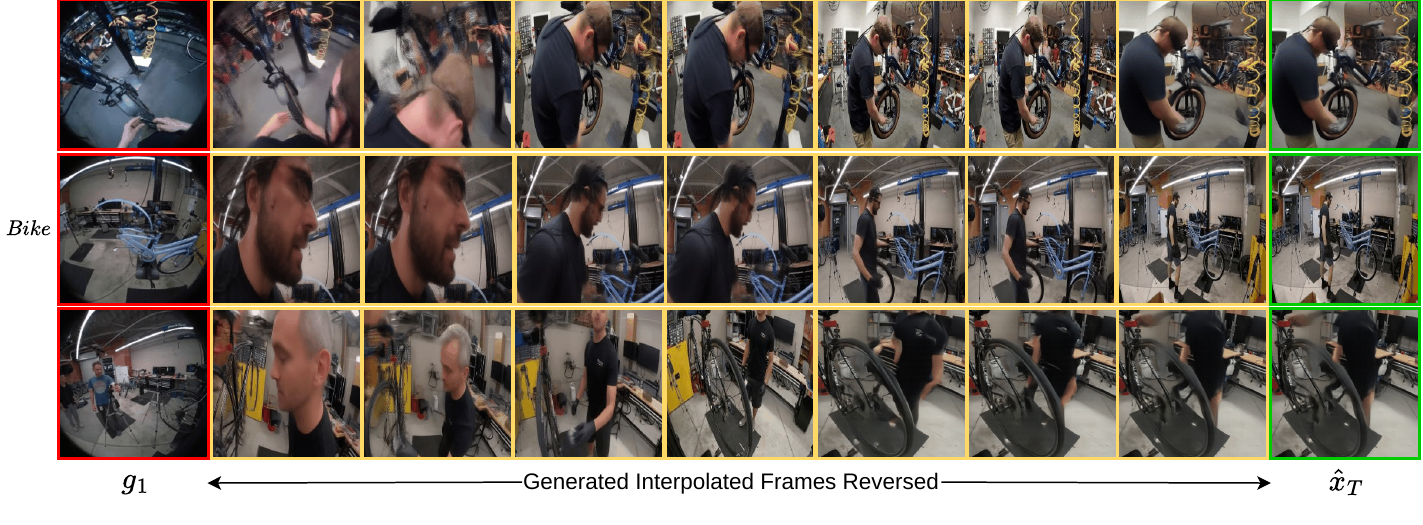}
  \caption{The model's ability to perform reverse interpolation from ego to exo viewpoints for the Ego2Exo task (\textit{bike} category).}
  \label{fig:ego2exo_bike}
  \vspace{-0.1in}
\end{figure}

We first present qualitative results demonstrating the model's ability to smoothly interpolate between exocentric and egocentric videos at inference time in Fig.~\ref{fig:interpolation}. We then provide additional qualitative examples illustrating the model's performance on the exocentric-to-egocentric cross-view video generation task in Fig.~\ref{fig:exo2ego}. Finally, we illustrate the extended Ego-to-Exo generation on the \textit{bike} subcategory in Fig.~\ref{fig:ego2exo_bike}.

\section{Failure Analysis}
We identify two main failure cases: (1) when multiple people are present, WFLF may occasionally interpolate to another person’s head; and (2) when the exocentric view does not fully cover the ego-view environment, e.g., when the exo camera is directly in front of the subject and the background behind the camera—required for the ego view—is not visible.

\section{Future Work}

While our experiments demonstrate that frame interpolation significantly improves Exo2Ego generation, there remains potential for further gains. For instance, increasing the number of backward diffusion steps when querying WFLF could produce higher-quality interpolated frames, though this would come at the cost of longer generation times. 

Our formulation divides each video into three segments and trains on two segments per iteration, resulting in a sequence length of $2T$. This design leads to faster convergence in practice. Although training the full $3T$ sequence could be more expressive, it potentially yields higher-quality results (it was not feasible within our computational budget). An interesting direction for future work is to jointly model all three segments within a single training iteration.

Regarding pose estimation, our current approach performs linear interpolation between the last exocentric frame and the first egocentric frame. This assumes a smooth and consistent progression between frames, which may not always hold in practice. Although interpolating poses separately from the video frames already improves performance compared to using no pose interpolation, a more integrated approach could be beneficial. One possibility is to leverage tools such as ViPE~\cite{huang2025vipe} to directly extract camera poses from the generated interpolated frames, aligning the pose trajectory more closely with the actual video content. However, this approach may introduce significant computational overhead.

Finally, our current fine-tuning is performed using only a subset of each category. Generating additional synthetic data for fine-tuning could further enhance model performance, particularly for categories with high variability in motion or interactions. Exploring these directions could lead to even more robust and realistic Exo2Ego video synthesis in future work.

\end{document}